\documentclass[conference]{IEEEtran}
\IEEEoverridecommandlockouts

\usepackage{amsmath,amssymb,amsfonts}
\usepackage{algorithmic}
\usepackage{graphicx}
\usepackage{textcomp}
\usepackage{xcolor}
\usepackage[numbers, sort&compress]{natbib}

\def\BibTeX{{\rm B\kern-.05em{\sc i\kern-.025em b}\kern-.08em
    T\kern-.1667em\lower.7ex\hbox{E}\kern-.125emX}}

\begin{document}

\title{Towards Reliable Medical Question Answering: Techniques and Challenges in Mitigating Hallucinations in Language Models}

\author{\IEEEauthorblockN{Duy Khoa Pham}
\IEEEauthorblockA{\textit{Department of Computing Technologies} \\
\textit{Swinburne University of Technology} \\
Melbourne, Australia \\
103515617@student.swin.edu.au}
\and
\IEEEauthorblockN{Bao Quoc Vo}
\IEEEauthorblockA{\textit{Department of Computing Technologies} \\
\textit{Swinburne University of Technology} \\
Melbourne, Australia \\
bvo@swin.edu.au}
}

\maketitle

\begin{abstract}
The rapid advancement of large language models (LLMs) has significantly impacted various domains, including healthcare and biomedicine. However, the phenomenon of hallucination, where LLMs generate outputs that deviate from factual accuracy or context, poses a critical challenge, especially in high-stakes domains. This paper conducts a scoping study of existing techniques for mitigating hallucinations in knowledge-based task in general and especially for medical domains. Key methods covered in the paper include Retrieval-Augmented Generation (RAG)-based techniques, iterative feedback loops, supervised fine-tuning, and prompt engineering. These techniques, while promising in general contexts, require further adaptation and optimization for the medical domain due to its unique demands for up-to-date, specialized knowledge and strict adherence to medical guidelines. Addressing these challenges is crucial for developing trustworthy AI systems that enhance clinical decision-making and patient safety as well as accuracy of biomedical scientific research.
\end{abstract}


\section{Introduction}
Large language models (LLMs) have opened up new avenues for their application in knowledge-intensive tasks including medical domain. However, hallucinations, which outputs that deviate from factual accuracy, present a critical challenge, especially in high-stakes domains like healthcare~\cite{PalMedhalt, sun2024trustllm}, where accuracy and reliability are paramount~\cite{puchert2023llmmaps}. Hallucinations in healthcare AI systems can lead to severe consequences, including incorrect clinical decision-making, delayed or improper treatment, and compromised patient safety~\cite{MORLEY2020113172}.

To address this critical issue, researchers have explored various techniques to mitigate hallucinations in knowledge-based tasks~\cite{tonmoy2024comprehensive, ji2023survey}, such as Retrieval-Augmented Generation (RAG)~\cite{lewis2020retrieval, mao-etal-2021-generation, peng2023check, vu2023freshllms, cao2023step, kang2023ever, rawte-etal-2023-troubling}, self-refinement via iterative feedback loops~\cite{ji-etal-2023-towards, si2022prompting, lei2023chain}, and supervised fine-tuning on factual data~\cite{tian2023fine}. While these techniques have shown promise in general domains, their effectiveness in the healthcare and biomedical contexts remains understudied, with only a few papers examining the issue. The medical domain presents unique challenges, including the need for up-to-date and specialized knowledge, strict adherence to established medical guidelines, and a deep understanding of the complex interplay between medical concepts and patient health outcomes. 

This paper presents a scoping study of hallucination mitigation techniques, assessing their effectiveness in improving accuracy and reliability for QA and summarization tasks, with a focus on adapting these methods to the medical domain. By examining the breadth of research in this critical area, we aim to enhance understanding of current LLM limitations in healthcare applications and provide insights into addressing hallucinations. This study contributes to the advancement of LLMs in healthcare AI, supporting the development of more trustworthy and reliable systems for clinical decision support and biomedical research. 

\section{Background}
\subsection{Definition of Hallucination}
In the context of language models, particularly those trained for tasks like question answering, translation, summarization, and dialogue systems, ``hallucination'' refers to the generation of content that is either unfaithful or irrelevant to the given input or context. Maynez et al.'s seminal paper~\cite{maynez-etal-2020-faithfulness} has categorized hallucination into two types: \emph{intrinsic hallucination} where the text generated by the model contradicts the facts or data provided in the input and \emph{extrinsic hallucination}, where the generation of information that cannot be verified or contradicted by the source input. Recently, Zhang et al.'s influential survey on hallucination in LLMs have provided a more granular categorization of hallucination within the context of LLMs ~\cite{zhang2023siren} with three primary types of hallucination:

\textit{Input Conflicting:} The model's response deviates from the user's input (including task instructions and input). 

\textit{Context Conflicting:} Generated content contradicts itself or previously generated content.

\textit{Fact Conflicting:} Generated content contradicts established world knowledge, posing significant challenges in real-world applications, especially in high-stakes domains like healthcare.

In medical domain, Ji et al.~\cite{ji-etal-2023-towards} classified problems with answers in medical QA tasks into three categories: 
(i) \textit{Fact Inconsistency}: Providing information inconsistent with established facts.
(ii) \textit{Query Inconsistency:} Providing information unrelated to the user's query.
(iii) \textit{Tangentiality:} Providing related but not directly answering the query.
Among these, fact inconsistency is the most challenging and common form of hallucination in medical domain~\cite{niu2024mitigating}.

\subsection{Causes of hallucination in Language Models}
Hallucinations in language models can critically undermine their utility, leading to outputs that are unfaithful or irrelevant to the input. This section explores the root causes of hallucination in language models. 

\subsubsection{Hallucination from Dataset}
The main reason for hallucination comes from the data source used to train the model. There are several factors for this issue:

\textit{Data Collection Flaws:} Hallucinations frequently originate from flaws in data collection methods. In large-scale dataset construction, heuristic methods are frequently used to pair real sentences or tables as source and target~\cite{lebret-etal-2016-neural}. However, this approach can introduce significant discrepancies. In particular, information in the target can be not verified by the source which is the factual data~\cite{parikh-etal-2020-totto}. This discrepancy means that the target reference may contain unsupported details, leading to hallucinations when the model generates text. Such mismatches between the source and target data introduce inconsistencies that models learn and replicate, resulting in hallucinated outputs.

\textit{Duplicate Data:} When duplicates are not effectively removed from the dataset, the model may overfit to these repeated instances. This leads to a model that is prone to generate repetitive or overly similar outputs across different inputs, contributing to hallucinations~\cite{lee-etal-2022-deduplicating}.

\textit{Inherent Task Misalignment:} Certain language generation tasks are inherently misaligned with the objective of factual accuracy due to their very nature. Tasks such as creative writing or generating speculative content are not necessarily bound by strict factual constraints. This fundamental lack of alignment with truthfulness can condition models to become more tolerant of deviating from factual accuracy, thereby increasing the likelihood of hallucinations. For example, models may learn from the behavior of some authors who state claims based on their feelings or personal perspectives rather than verifiable facts. This unverified input data teaches the model that such subjective assertions can be considered ``correct''. Consequently, the model may learn these behaviors, potentially causing it to extrinsically hallucinate.

\subsubsection{Hallucination from Model Training and Inference}
Even with minimal factual divergence in the training data, hallucinations can still occur during generation~\cite{parikh-etal-2020-totto}. This suggests that other factors related to the model's training or the learning algorithm itself contribute to the production of hallucination:

\textit{Encoder Limitations:} If the encoder component of a model does not effectively understand or encode the nuances of the input data, it may fail to capture important details that are crucial for generating accurate output~\cite{parikh-etal-2020-totto}. This could lead to the generation of irrelevant or incorrect information (hallucinations) compared to the input~\cite{aralikatte-etal-2021-focus, Feng_Xie_Gu_Shao_Zhang_Yang_Yu_2020}.

\textit{Decoding Problems:} There are primarily two reasons for hallucinations in language models: decoder misalignment and flaws in decoding strategies. Regarding decoder misalignment, the decoder can attend to the wrong part of the encoded input source, leading to erroneous generation~\cite{tian2019sticking}. This wrong association results in generated text with facts mixed up between similar entities~\cite{dziri-etal-2021-neural}. As for decoding strategy flaws, the design of the decoding strategy itself can contribute to hallucinations. Decoding strategies that improve generation diversity, such as top-k sampling, are positively correlated with increased hallucination~\cite{dziri-etal-2021-neural}. The deliberate introduction of ``randomness'' by sampling from the top-k samples instead of choosing the most probable token may increase the unexpected nature of the generation, leading to a higher chance of containing hallucinated content.

\textit{Parametric Knowledge Bias:} Large language models accumulate a broad base of ``knowledge'' encoded within their parameters during the pre-training process on massive datasets. While this parametric knowledge aids performance on downstream tasks, it also contributes to hallucinatory generations. The powerful generalisation ability of large pre-trained models for natural language generation comes at the cost of prioritizing their parametric knowledge over strictly adhering to provided input prompts~\cite{longpre-etal-2021-entity}. In other words, models favoring outputs aligned with their internal knowledge base rather than the specific input information are more prone to hallucinating excess or inconsistent content.

\subsubsection{Snowball Hallucination}
One phenomenon that can contribute to hallucinations in language models is known as ``snowball hallucination'' which is thoroughly examined by Zhang et al.~\cite{zhang2023language}. The paper investigates how initial minor inaccuracies in language models can lead to significant compounded errors, relating to discussion on mitigating hallucinations in QA tasks. 
The paper questions whether knowledge gaps are the sole source of hallucination in LMs. The hypothesis is that LMs are prone to ``hallucination snowballing'', a phenomenon where initial minor inaccuracies lead to significant compounded errors. This occurs because instruction-tuned LMs often commit to an answer in the first token and continue to provide a supportive explanation, regardless of the correctness of the initial answer. The paper identifies two key findings:
\begin{enumerate}
    \item \textit{Initial Committal:} LMs, especially those tuned for instructions, tend to generate an answer before providing an explanation due to the design of instruction data. This design leads models to commit to a binary answer (e.g., Yes/No), and the subsequent explanation supports this answer to maintain coherence, even if the initial answer is wrong. 
    
    \item \textit{Inherently Sequential Processing:} Transformer-based models process inputs and generate outputs one token at a time, making them ill-suited for tasks requiring multi-step reasoning. This limitation forces the model to provide answers to complex questions in a single step, increasing the likelihood of generating an incorrect answer followed by an incorrect justification.

\end{enumerate}

The study designed three datasets—Primality Test, Senator Search, and Graph Connectivity—to induce hallucination snowballing with yes/no questions that cannot be answered in a single step. The findings revealed that ChatGPT (2022) and GPT-4 (2023) gave correct yes/no answers 39.87\% and 16.6\% of the time, respectively. Despite their ability to detect incorrect claims (67.37\% for ChatGPT and 87.03\% for GPT-4), they often still generated wrong answers, demonstrating the existence of hallucination snowballing~\cite{zhang2023language}. When the prompt included ``Let's think step by step,'' error rates were significantly reduced, showing that encouraging step-by-step reasoning helps mitigate snowball hallucinations.

\section{Approach}
\subsection{Research Questions}
To address the critical issue of hallucinations in large language models for knowledge-based tasks in the medical domain, this study aims to systematically investigate the following research questions:
\begin{enumerate}
    \item[RQ1:] How effective are current hallucination mitigation techniques for knowledge-based tasks such as QA and summarization? 
    \item[RQ2:] How effective are hallucination mitigation techniques in improving the accuracy and reliability of medical QA and summarization?
\end{enumerate}

\subsection{Search Strategy}
To conduct this scoping study, we utilize both manual and automated search techniques to ensure a comprehensive coverage of relevant literature.

\subsubsection{Manual Search}
The manual search process involves identifying relevant papers based on keywords derived from the research questions. These initial papers serve as a foundation for extracting search strings, which are subsequently used in the automated search process.

\subsubsection{Search String Formation and Automated Search}
The search strings are constructed by extracting relevant terms from the manually identified papers. These search strings aim to encompass keywords pertinent to the research questions. Utilizing the formulated search strings, an automated search is conducted across various databases and scholarly repositories. This automated search process is followed by a snowballing approach, which involves examining the reference lists and citations of the initially retrieved papers to identify additional relevant literature.

\subsubsection{Literature Review and Filtering}
After obtaining the search results, the next step involves a thorough literature review process. This process involves extracting and documenting relevant information from each paper, such as the publication year, journal or conference, authors, keywords, mitigation techniques employed, problem addressed, key contributions, technical metrics or criteria used, and any identified issues or limitations. Subsequently, the collected papers are filtered to align with the specific research questions of the study.

\subsubsection{Inclusion of Recent Literature}
To ensure that the study incorporates the latest advancements and developments in the field, the search strategy also involves updating the literature review with recently published papers from open-access repositories such as arXiv. This step ensures that the systematic literature review remains current and captures the most recent research findings and perspectives.

\section{RQ1: How effective are current hallucination mitigation techniques for knowledge-based tasks such as QA and summarization?}

\subsection{Comprehensive Literature Reviews of Hallucination Mitigation Techniques}
Many papers have focused on exploring and analysing various techniques for mitigating hallucination in LLMs. Among them, Tonmoy et al.'s survey~\cite{tonmoy2024comprehensive} stands out by comprehensively categorizing 32 different techniques aimed at hallucination mitigation. This systematic survey introduces a taxonomy of hallucination mitigation techniques for LLMs by categorizing them into distinct phases and methodologies such as Retrieval-Augmented Generation, Prompt Engineering, and Developing Models. Furthermore, the taxonomy illustrates the temporal aspects of RAG techniques, such as ``Before Generation'', ``During Generation'', ``After Generation'', and ``End-to-End'', which provides insights into the stages at which these techniques are employed during the generation process. Inspired by this taxonomy, we further explore these techniques with a focus on knowledge-based tasks like QA or summarization, especially for medical domain.

Another significant contribution to the literature by Ji et al.~\cite{ji2023survey} provides a comprehensive overview of the different types of hallucinations, categorized as intrinsic and extrinsic based on previous work. It delves into the causes of these hallucinations and explores various strategies to mitigate them. This paper serves as a foundational reference for understanding the diverse types of hallucinations and their implications in natural language generation. The insights gained from this survey are crucial for developing targeted mitigation strategies, especially in applications within the medical domain, where the accuracy and reliability of information are of paramount importance.

\subsection{Retrieval-Augmented Generation (RAG)}
RAG has emerged as a promising technique to enhance the performance and reliability of LLMs by incorporating external knowledge during the text generation process. This approach addresses the limitations of LLMs in generating accurate and contextually relevant information, particularly in knowledge-intensive tasks such as medical question answering (QA) and summarization. The following sections provide a overview of RAG and discuss various studies categorized by the phase of RAG they try to enhance: before generation, during generation, after generation, and end-to-end training.

\subsubsection{Comprehensive Overview of RAG}
RAG involves the integration of retrieval mechanisms with generative models to improve the quality of generated text. By fetching relevant information from external sources, RAG systems can ground their responses in factual data, thereby reducing the incidence of hallucinations. A recent survey by Hu and Lu~\cite{hu2024rag} presents an overview on Retrieval-Augmented Language Models (RALMs) which incorporate external information retrieval to enhance the capabilities of large language models across various NLP tasks. It provides a comprehensive survey of both RAG and RAU methods, detailing how these approaches improve the performance and reliability of language models in processing natural language. It gives detailed summary about aspect of RALMs such as: definition, LM, retriever, enhancement, application and so on.

\subsubsection{RAG Before Generation}
Peng et al.~\cite{peng2023check} explore the concept of generation-augmented retrieval for open-domain question answering (QA). This approach utilizes retrieved knowledge during the generation process, aiming to enhance the accuracy and reliability of the outputs produced by large language models. 

In~\cite{vu2023freshllms}, Vu et al. investigate the potential of refreshing and updating the knowledge base of large language models with search engine data. This process aims to significantly improve the models' accuracy and relevance. A key contribution of this study is the development of a practical approach for keeping medical models up-to-date with the latest research and clinical guidelines, which could potentially reduce the incidence of hallucinations in these models. 

\subsubsection{RAG During Generation}
Cao et al.~\cite{cao2023step} propose a novel multi-stage question decomposition framework that leverages RAG to systematically refine the information retrieval process. The key innovation lies in ensuring that each stage of the question decomposition focuses on extracting only the most relevant and reliable information from external sources. This approach aims to constrain the reasoning process, preventing the inclusion of erroneous or irrelevant data, which is particularly critical for reducing hallucinations in QA systems. The stringent accuracy requirements in healthcare necessitate precise and trustworthy information, making this method a promising solution for mitigating hallucinations and enhancing the reliability of AI-powered decision support systems in healthcare. 

Kang et al.~\cite{kang2023ever} introduce a novel approach called EVER, which goes beyond traditional RAG methods. EVER employs a real-time validation and rectification process to address inaccuracies during the generation phase, aiming to enhance the reliability of language model outputs. 

In~\cite{mao-etal-2021-generation}, Mao et al. investigate the integration of generative models with retrieval processes to improve the depth and relevance of information accessed during question-answering tasks. The key contribution of this approach is that by augmenting retrieval with generation, it not only expands the pool of available information for answering questions but also ensures that the retrieved information is highly contextualized, thereby reducing the likelihood of hallucinations caused by out-of-context data. 

\subsubsection{RAG After Generation}
Rawte et al.~\cite{rawte-etal-2023-troubling} introduce a unique framework for quantifying hallucinations in language models. It proposes both qualitative and quantitative methods for assessing hallucinations, emphasizing the need for a nuanced understanding of different hallucination types. Specifically, it categorizes hallucinations into intrinsic and extrinsic types, and further divides them into mild, moderate, and severe categories, each with specific implications and remediation strategies. 

\subsubsection{RAG End-to-End}
In~\cite{lewis2020retrieval}, Lewis et al. investigate the integration of parametric memory (model knowledge) with non-parametric memory (external documents) using a RAG-based approach for knowledge-intensive NLP tasks like open-domain question answering. This method employs end-to-end training, where a retriever fetches relevant documents from a large external dataset, and a generator produces the final output based on both the input query and the retrieved documents, designed to work seamlessly together. 

\subsubsection{Optimizing Fidelity and Utilization in RAG Models}
Wu et al.~\cite{wu2024faithful} analyze the effectiveness of RAG models in maintaining fidelity to source information, particularly examining how the internal biases of LLMs can affect the accuracy of retrieved data. It provides a quantitative analysis of the conflicts between the generative tendencies of LLMs and the corrective capabilities of RAG. The key contribution is offering insights into optimizing RAG to prioritize factual accuracy, which is essential for medical applications where misinformation can have severe consequences. The study emphasizes the need to balance the generative and retrieval aspects to enhance model reliability in sensitive healthcare environments.

To address the aforementioned concerns, Labruna et al.~\cite{labruna2024retrieve} introduce ADAPT-LLM, a model trained to dynamically decide whether to use its parametric knowledge or seek external information via information retrieval (IR). The design rationale describes the necessity for LLMs to identify when to rely on stored knowledge versus when to retrieve additional information. ADAPT-LLM is trained to generate a special token \texttt{<RET>} signaling the need for external data to improve question-answering capabilities. A key contribution is to demonstrate the feasibility of teaching LLMs to efficiently use external sources only when necessary.

\subsection{Other Methods}
Si et al.~\cite{si2022prompting} focus on improving GPT-3’s performance through targeted prompting strategies, this study explores how effectively tailored prompts can reduce hallucinations and increase reliability across various tasks. By refining the prompt engineering process, the researchers were able to significantly enhance the model's response quality, which is particularly vital in high-stakes environments where precise information is paramount. This method provides a scalable way to improve the utility and accuracy of language models in real-world applications, ensuring that they produce dependable and contextually appropriate outputs.

Lei et al.~\cite{lei2023chain} introduce a method that utilizes a chain of natural language inference tasks to verify each step of text generation, ensuring that every link in the chain is grounded in factual accuracy. This systematic verification significantly reduces the incidence of ungrounded hallucinations, where the model generates baseless claims. Such an approach is indispensable in domains where the veracity of information is crucial, such as in medical or legal applications, providing a robust framework for ensuring the reliability of automated text generation systems.

The UPRISE study~\cite{cheng-etal-2023-uprise} enhances zero-shot task performance by utilizing dynamically retrieved prompts that guide the language model's generation process. This innovative approach to prompt engineering uses a retrieval system to select the most effective prompts dynamically, improving the model's performance and reducing the risk of hallucinatory outputs. By aligning prompt selection with specific task requirements, this method significantly bolsters the model's ability to generate accurate and contextually relevant responses, showcasing a promising direction for enhancing the adaptability and effectiveness of language models in diverse applications.

In the Inference-Time Intervention study~\cite{li2024inference}, a novel decoding strategy is proposed that involves real-time interventions during the model’s inference process, guiding it towards more truthful and accurate outputs. This method is crucial for applications where precision is necessary, as it ensures the integrity of the information generated by the model. By focusing on veracity at the time of inference, this approach addresses the prevalent issue of hallucinations, enhancing the trustworthiness and reliability of responses in sectors such as healthcare and finance.

The R-Tuning research~\cite{zhang2023r} explores techniques to train language models to recognize their limitations and refuse to answer when they lack sufficient information. By training models to refuse to respond rather than fabricating answers, the study aims to enhance the trustworthiness of AI systems, particularly in medical scenarios where incorrect information can result in detrimental outcomes. This refusal skill is a critical development towards creating more reliable and ethical AI systems, ensuring that they contribute positively in high-stakes environments.

Large Language Model unlearning by Yao et al.~\cite{yao2023large} is an emerging technique aimed at addressing and rectifying specific undesirable behaviors in LLMs by effectively "forgetting" or suppressing them. This innovative approach offers a promising solution for mitigating hallucinations by enabling targeted refinement of model outputs without necessitating comprehensive retraining. Unlearning stands out for its efficiency, requiring primarily negative examples, which makes it a cost-effective alternative to more resource-intensive methods like reinforcement learning from human feedback (RLHF).

\subsection{Insights and Future Directions}
\textit{Importance of Data Quality and Source Authority for RAG:} The reliability of external data sources is paramount for minimizing hallucinations for RAG. Models need to prioritize authoritative sources to ensure the generated content is accurate. Training models to recognize and validate the authority of these sources can significantly reduce the incidence of hallucinations. High-quality data should be clean, up-to-date, and relevant to the specific domain, particularly in sensitive areas like biomedicine where accuracy is critical. Future research should focus on developing more sophisticated retrieval mechanisms aiming at better prioritize authoritative sources and high-quality data.

\textit{Dynamic Retrieval Decisions:} Models that can dynamically decide when to use internal knowledge versus external retrieval show potential promise. This decision-making process is crucial for maintaining the accuracy and relevance of the generated responses. Further exploration of techniques like ADAPT-LLM~\cite{labruna2024retrieve}, which train models to signal the need for external data dynamically, can improve the model's ability to provide accurate answers by leveraging both internal and external knowledge effectively.

\textit{Real-Time Validation and Rectification:} Real-time updates to the model’s knowledge base can significantly enhance its accuracy. Techniques that enable models to refresh their knowledge with the latest information are crucial, especially in fields like biomedicine where new research and guidelines are constantly emerging. Techniques like EVER~\cite{kang2023ever} employ real-time checks to correct inaccuracies during the generation phase. More research is needed to develop efficient real-time validation mechanisms without significantly impacting model performance.

\textit{Conflict Resolution in Pre-Trained Models:} Resolving conflicts between pre-trained model knowledge and retrieved information remains a challenge. Effective strategies are needed to resolve these conflicts, ensuring that the model can accurately decide when to rely on internal knowledge and when to seek external validation. Future work should explore methods to effectively integrate and prioritize different knowledge sources.

\textit{Iterative Feedback and Self-Refinement:} Self-refinement techniques involving iterative feedback loops enable models to continuously evaluate and improve their outputs. A promising recent approach, such as Meta-Rewarding LLMs~\cite{wu2024meta}, employs a three-role system—actor, judge, and meta-judge—which shows significant potential and may serve as an alternative to RLHF.

\textit{Ensemble Learning:} Ensemble methods, such as combining multiple models, can help mitigate individual biases and improve overall accuracy. Techniques like the LLM-Synergy~\cite{yang2023one} leverage ensemble learning to enhance performance on medical QA systems. By combining different models and leveraging the strengths of each, ensemble methods can significantly reduce hallucinations in the medical domain. Researching deeper into ensemble learning for hallucination reduction involves exploring how to effectively integrate various models, each excelling in different aspects, to produce more reliable and accurate outputs.

\section{RQ2: How effective are hallucination mitigation techniques in improving the accuracy and reliability of medical QA and summarization?}

\subsection{Benchmark and Literature Review Relating to Medical Domain}
BioMedLM~\cite{bolton2024biomedlm} is a comprehensive language model consisting of 2.7 billion parameters, trained exclusively on biomedical texts from PubMed. This extensive training enables BioMedLM to effectively address the specific needs of the medical domain, making it a robust tool for researchers and practitioners. A key contribution of BioMedLM is its availability on platforms like Hugging Face, which makes it accessible and suitable for various biomedical research applications, providing a valuable resource for advancing medical research and applications.

Wang et al.~\cite{wang2023pre} review the advancements in the use of pre-trained language models for biomedical applications, highlighting progress made while identifying existing gaps in current methodologies. The survey offers a broad overview of the landscape, providing essential context for applying specific hallucination mitigation techniques effectively. This comprehensive review is crucial for developing strategies to enhance the accuracy and reliability of biomedical language models, thus improving their practical utility in the medical field.

Med-HALT~\cite{PalMedhalt} introduces a new benchmark dataset designed to evaluate medical hallucinations in large language models, covering a wide range of medical knowledge and reasoning/memory-based tests. The dataset aims to address the need for domain-specific hallucination evaluation in the medical field, as existing solutions are often too general. By reducing the reliance on human annotation and assessing both hallucinations and factual accuracy, Med-HALT provides a more efficient and targeted evaluation method tailored to medical contexts.

This literature review by Wubineh et al.~\cite{wubineh2023exploring} systematically examines the ethical, social, privacy, and technological aspects of adopting artificial intelligence in healthcare. It identifies critical challenges, including data privacy concerns, bias mitigation needs, a lack of awareness about AI's benefits and limitations, transparency issues with algorithms, and potential over-reliance on AI. Additionally, the review highlights opportunities such as enhanced decision support, technological advancements in diagnostics and drug development, and improved patient monitoring through AI technologies. These insights are essential for understanding the complex landscape of AI implementation in healthcare.

Gao et al.~\cite{gao2023examining} study explores the use of ChatGPT (GPT-3.5 model) for identifying correct and incorrect drug-disease associations through prompting. The findings indicate that ChatGPT achieved an accuracy range of 74.6-83.5\% for true associations and 96.2-97.6\% for false ones. Providing additional disease context through prompts improved the accuracy of true associations but slightly decreased the accuracy for false associations. While ChatGPT performed better at ruling out incorrect associations, limitations include hallucinations and a lack of the latest biomedical knowledge, highlighting areas for improvement in biomedical information retrieval.

Pushpanathan et al.~\cite{pushpanathan2023popular} study evaluation for the performance of ChatGPT-3.5, ChatGPT-4.0, and Google Bard in accurately and comprehensively answering 37 common questions about eye and vision symptoms. The results show that ChatGPT-4.0 outperformed ChatGPT-3.5 and Google Bard in terms of accuracy, with all three models demonstrating high comprehensiveness scores for good responses. Notably, ChatGPT-3.5 frequently disclaimed its responses, while ChatGPT-4.0 and Bard often asserted accuracy even when incorrect, leading to hallucinations. This study underscores that although current LLMs have significant capabilities, they cannot fully replace the expertise of ophthalmologists.

\subsection{Hallucianation Mitigation Techniques in Medical Domain}
Ji et al.~\cite{ji-etal-2023-towards} utilize iterative feedback loops to refine the accuracy of medical QA systems, focusing on reducing hallucinations through self-assessment and adjustment. This approach demonstrates a practical application of self-refinement techniques in the medical domain, where accuracy is critical for patient outcomes. By continuously evaluating and adjusting the model's outputs, the technique aims to enhance the reliability and factual correctness of generated responses.

Another study by Tian et al.~\cite{tian2023fine} explores supervised fine-tuning methods to improve the factual accuracy of language models, particularly for biography generation and medical QA tasks. The study employs automated factuality preference ranking and direct preference optimization (DPO) to train models without human labels, using the FactScore metric to evaluate the results. This fine-tuning process ensures that the models generate more factual and reliable content, which is vital for applications in the medical field.

The paper by Ahmad et al.~\cite{ahmad2023creating} discusses various methods for evaluating and measuring the trustworthiness of LLMs in healthcare, including both human and automated evaluation techniques. It emphasizes the need for robust regulatory and evaluation frameworks to ensure the reliability of AI systems in sensitive domains. The insights provided are crucial for developing and deploying AI technologies that adhere to principles of transparency, non-biased, and ethical considerations in healthcare.

Yang et al.~\cite{yang2023one} propose the LLM-Synergy pipeline, which combines multiple large language models using ensemble methods like Boosting-based Weighted Majority Vote and Cluster-based Dynamic Model Selection. This approach improves performance by mitigating individual model biases and weaknesses, offering scalability, flexibility, and computational efficiency. The ensemble method has shown superior performance on datasets like MedMCQA, PubMedQA, and MedQA-USMLE, making it a promising solution for enhancing the accuracy of medical QA systems.

Xu et al.~\cite{xu2024bmretriever} introduce a series of dense retrievers aimed at improving biomedical retrieval performance. The development of effective biomedical retrieval models faces challenges due to limited publicly annotated data and computational resources. BMRETRIEVER leverages a two-stage training process to address these challenges. The first stage involves unsupervised contrastive pre-training on large biomedical corpora, utilizing extensive and diverse biomedical data. This stage employs techniques for constructing positive and negative query-passage pairs to enhance the model's ability to discern relevant information from irrelevant data. The second stage involves instruction fine-tuning using labeled datasets, where high-quality labeled datasets and synthetic data generation are used for fine-tuning. This process significantly impacts the model's understanding and performance in biomedical retrieval tasks. Extensive experiments across multiple tasks and datasets demonstrate BMRETRIEVER’s efficiency and effectiveness, achieving superior performance compared to larger models with fewer parameters. This approach marks a significant advancement in the field of biomedical text retrieval, offering a robust solution to the challenges of limited annotated data and computational constraints.

\subsection{Insights and Future Directions}
\textit{Medical Domain-Specific Challenges:} The medical domain presents unique challenges including the need for up-to-date and specialized knowledge, strict adherence to established medical guidelines, and a deep understanding of complex medical concepts. Hallucinations in this context can have severe consequences, including incorrect clinical decisions and compromised patient safety, making it imperative to develop robust mitigation techniques.

\textit{Evaluation Metrics and Benchmarks:} Developing robust evaluation metrics and benchmarks specific to the medical domain is essential. These tools help assess the effectiveness of different hallucination mitigation techniques and guide improvements. Further research should explore more medical benchmarks like Med-HALT~\cite{PalMedhalt}. Such domain-specific benchmarks evaluate medical hallucinations, reducing reliance on human annotation while focusing on factual accuracy and reliability.

\textit{Effectiveness of RAG in Medical QA:} RAG has shown promise in improving the accuracy and reliability of medical QA systems by grounding generated content in authoritative external sources. However, the effectiveness of RAG depends heavily on the quality and relevance of the retrieved data. Techniques like ``FreshLLMs''~\cite{vu2023freshllms} that use search engine augmentation to keep models updated with the latest research can significantly enhance the performance of medical QA systems.

\textit{Open Domain vs. Specific Domain Training:} The trade-offs between fine-tuning open-domain models for specific tasks and training domain-specific models from scratch require further investigation. Open-Domain models can offer flexibility and adaptability, allowing them to be fine-tuned for various tasks. Tian et al.~\cite{tian2023fine} demonstrated the effectiveness of this approach in reducing hallucinations in the medical domain. Conversely, domain-specific models can provide higher initial accuracy for specialized tasks like BioMedLM~\cite{bolton2024biomedlm} but may lack flexibility and adaptability to new data or tasks. Understanding the trade-offs between these approaches is essential for optimizing the training and deployment of models in the medical domain.

\textit{Iterative Feedback and Continuous Improvement:} Iterative feedback loops and real-time validation techniques are crucial for maintaining the accuracy and reliability of medical QA systems. Self-refinement methods, as demonstrated in studies by Ji et al.~\cite{ji-etal-2023-towards} or Niu et al.~\cite{niu2024mitigating} show significant potential in enhancing the reliability of generated responses. Continuous improvement and adaptation are necessary to keep up with the rapidly evolving medical knowledge, ensuring that the models remain relevant and accurate.

\section{Conclusion}
In conclusion, this scoping study has highlighted the essence of hallucination mitigation in LLMs, particularly for knowledge-intensive tasks in medical domain. While techniques like RAG, self-refinement and unlearning show promise, significant challenges remain. These include the need for high-quality, domain-specific data sources, robust evaluation metrics, and methods to handle the unique complexities of medical information. Future research should focus on refining these techniques, developing more sophisticated real-time validation mechanisms, and addressing the ethical implications of AI in healthcare. Ultimately, the goal is to create AI systems that can reliably support clinical decision-making and enhance patient care, while maintaining the highest standards of accuracy and safety.
\bibliographystyle{plain}
\bibliography{references}
\end{document}